\documentclass[conference]{IEEEtran}
\IEEEoverridecommandlockouts
\usepackage{cite}
\usepackage{amsmath,amssymb,amsfonts}
\usepackage{graphicx}
\usepackage{textcomp}
\usepackage{xcolor}
\def\BibTeX{{\rm B\kern-.05em{\sc i\kern-.025em b}\kern-.08em
    T\kern-.1667em\lower.7ex\hbox{E}\kern-.125emX}}

\usepackage{adjustbox}
\usepackage{balance}
\usepackage{comment}
\usepackage{tabularx}
\usepackage{nth}
\usepackage{tikz}
\usepackage{varwidth}
\usepackage{multirow}

\usepackage{algorithm}
\usepackage{algpseudocode}

\usepackage{booktabs}

\usepackage{caption}
\usepackage{footnote}
\usepackage[flushleft]{threeparttable}

\usepackage{amssymb}

\usepackage{subcaption}
\usepackage{tcolorbox}

\usepackage{subcaption}
\makeatletter
\def\BState{\State\hskip-\ALG@thistlm}
\makeatother

\makeatletter
\renewcommand*\env@matrix[1][*\c@MaxMatrixCols c]{%
	\hskip -\arraycolsep
	\let\@ifnextchar\new@ifnextchar
	\array{#1}}
\makeatother

\begin{document}

\title{Personalized Model-Based Design of Human Centric AI enabled CPS for Long term usage
}

\author{\IEEEauthorblockN{Bernard Ngabonziza, Ayan Banerjee, Sandeep K.S. Gupta}
	\IEEEauthorblockA{
		\textit{Arizona State University}\\
		\{bngabonz, abanerj3, sandeep.gupta\}@asu.edu}

}

\maketitle

\begin{abstract}

Human centric critical systems are increasingly involving artificial intelligence to enable knowledge extraction from sensor collected data. Examples include medical monitoring and control systems, gesture based human computer interaction systems, and autonomous cars. Such systems are intended to operate for a long term potentially for a lifetime in many scenarios such as closed loop blood glucose control for Type 1 diabetics, self-driving cars, and monitoting systems for stroke diagnosis, and rehabilitation. Long term operation of such AI enabled human centric applications can expose them to corner cases for which their operation is may be uncertain. This can be due to many reasons such as inherent flaws in the design, limited resources for testing, inherent computational limitations of the testing methodology, or unknown use cases resulting from human interaction with the system. Such untested corner cases or cases for which the system performance is uncertain can lead to violations in the safety, sustainability, and security requirements of the system. In this paper, we analyze the existing techniques for safety, sustainability, and security analysis of an AI enabled human centric control system and discuss their limitations for testing the system for long term use in practice. We then propose personalized model based solutions for potentially eliminating such limitations.  

\end{abstract}


\maketitle

\IEEEdisplaynontitleabstractindextext

%
\IEEEpeerreviewmaketitle

\section{Introduction}
Human centric monitoring and feedback systems are increasingly being used in practical settings reaching a significant user base. Examples include autonomous driver assist systems, wearable sensor based health monitoring systems, gesture based communication interfaces, and medical control systems such as closed loop blood glucose control systems for Type 1 Diabetic subjects. The primary characteristics of these applications are they are cyber-physical systems. This is because they involve closed loop collaboration between the human user and the machine. In addition most of these systems have some component that includes an artificial intelligence mechanism. Moreover many of such applications are to be used in critical scenarios for long term. Consider the example of a closed loop glucose control system also known as Artificial Pancreas (AP). The AP system is a closed loop system with a continuous glucose monitor (CGM) sensor sensing glucose levels from the tissue fluid and sending it to an infusion pump. This pump has a control software that uses an adaptive intelligent algorithm to first predict the blood glucose level 30 mins in the future and compute the current infusion rate to keep the future blood glucose level within normal limits. The insulin is then infused by the pump at a steady rate. This system is a cyber-physical system and is enabled by an AI software. It is also intended to be used by people suffering from Type 1 diabetes on a regular basis 24/7 for a  long period of time, typically years. 

An important characteristics of such systems is that since they are used in critical applications often these are not fully automatic. This means that certain components of the system requires input from the human user. For example, in case of the AP system, the human user is required to provide information about meal timings, carbohydrate content of the food (best guess), and the amount of insulin to be taken to offset the glucose rise due to the meal. In another example, such as autonomous cars, the automatic braking system can be enabled or disabled based on human input. 

Critical nature of usage scenarios for such systems require rigorous testing before deployment. The broad objectives are safety, sustainability, and security of the system. A mixed approach towards testing is undertaken, which includes model based verification and experimental validation. However, such techniques are often not comprehensive due to several reasons including: a) lack of resources, computational, storage, and time, b) inherent algorithmic complexity that often makes the verification problem intractable, and c) occurrence of unknown operating situations that only exhibit during deployment and are not taken into account during testing. Moreover, since the verification problem is often intractable, approximations are used to determine system behavior for a large set of initial conditions. Such approximations may result in uncertainty of the verification result for a certain operating configuration of the system. When a system is used for a long term in real life deployments, configurations that are untested or for which the verification result is uncertain can occur in critical scenarios leading to potentially fatal safety, security, or sustainability violations. The following examples illustrate our argument:

\noindent{\bf Vignette 1: Post prandial hyperglycemia in AP systems -} The AP system was introduced for Type 1 diabetic subjects to control increase in blood glucose levels specifically after meal. However, a side effect of using AP system is that while controlling high blood glucose levels it can induce hypoglycemia if too much insulin is infused in the body, leading to potential safety risks that can be fatal. Current controller designs have algorithms to significantly reduce hypoglycemia. Controlled experimental studies have shown that AP systems can reduce hypoglycemia while controlling post prandial blood glucose levels. However, several post market studies including analysis performed by the authors show that subjects spend significant amount of time in post prandial hyper-glycemia. Deeper analysis shows that the time in hyperglycemia is directly proportional to meal intake suggesting that the problem is related to use cases that are unseen during test time.

\noindent{\bf Vignette 2: Fatal safety violation in autonomous cars -} A recent incident in Phoenix Arizona with Uber autonomous vehicle lead to a fatal accident. The NTSB report on the accident suggests that the Uber backup driver did not configure the Uber vehicle for automatic braking in case of emergencies. In addition the Uber object recognition system recognized the person as an unknown object, a car and then a bicycle and never a pedestrian with varied predicted future travel paths. Such a fatal error was not caught during test time. The primary reason for such an error is not well understood yet but potential reasons can be uncertainty in verification results, or lack of incorporation of the use case during test time.

\noindent{\bf Vignette 3: Data provenance in wearable sensors -} In several practical deployments of wearable sensor based monitoring systems which are connected to a cloud service verifying the source of the data is an important problem. Such verification is not carried out typically for personal wearable monitoring systems that do not have a cloud interface. The inherent assumption is that the subject using the system is less likely to subvert the system. However, when a wearable monitoring system is used for the purpose of managing a chronic health problem which has a significant behavioral aspect such as type 2 diabetes, addiction, or obesity, then it is possible that the user of the system can attempt to subvert the source of the data to influence the result of data analysis. Data provenance has not been studied under the attack model where the users themselves attempt to obfuscate the data source.

In this paper, we first focus on the unique challenges brought about by the long term practical deployments of AI enabled cyber human systems to guarantee their safety, security, and sustainability. We take examples from the medical domain such as closed loop blood glucose control system or artificial pancreas, cardiac monitoring systems, and body sensor networks in general to contextualize and evaluate the impact of these challenges. Finally, we propose a unique solution direction towards solving safety, security and sustainability challenges by incorporating tighter coupling of the human with the machine through the development and usage of personalized models.

\section{System Model}

Our system consists of several hardware devices and design layers. These layers consists of  
1) \textit{perception} (which gathers information and influence the action of the environment 
through sensing and actuation.), 2) \textit{network} (responsible for the communication between 
different devices), 3) \textit{service} (which provides various services, such as data abstraction
or running security protocols for the other three layers) and 4) \textit{application} 
( for interaction between the patient,caregiver, and the system itself) layers. 

These wearable medical systems contain a number of diverse, low-cost, wireless embedded 
{\em sensors} and a few {\em actuator} which together form a {\em distributed wireless network} 
around the patient  \cite{Kermani}. The sensors continuously monitor various physiological signals 
from the patient and wirelessly forward them to a {\em base station/sink} entity, usually implemented on a smartphone;  which is responsible for managing therapies, using the actuators present. The sink is also responsible for complex visualization, storage and forwarding the patient data to a medical cloud. 

\subsection{Sensors/Actuators}

The perception layer is responsible for influencing the environment and gathering information 
from through sensing and actuation. The main objective of the perception layer is to gain information from the environment and trigger some actions in response to the perceived information using sensors and actuators, respectively. These end devices are also called as \textit{nodes} in IoT-based systems.


\subsection{Mobile Devices}
Sensors can stream data to mobile phone via Blue-tooth in real-time. The mobile phones can host a set of control algorithms that determine the actuation inputs or they can merely act as a data forwarder to the cloud. In some applications such as the Medtronic closed loop blood glucose control systems the controller and the actuator is combined into a single device.  

\subsection{Cloud Server}
The cloud server is data storage and computation hub. It is not only used as a computational and storage resource but also used as a knowledge resource in many AI applications. The large scale data repositories that are available in cloud hosted systems can be used to aide the development of predictive models.

\subsection{Safety, Security, and Sustainability Requirements}
In this paper, we make a focused discussion on the safety, security, and sustainability requirements. By safety, we mean the safety of the human in the system. Typically safety is either expressed as a constraint on some continuous parameters in the system. With such a definition safety violations becomes a threshold crossing event. In addition to this definition safety violations can also be random events. 

A system is considered sustainable, if the cost of resources required to continually run the system remains within a limit set forth in design time. As such sustainability can have several manifestations, however, in this paper we will only focus on sustainability manifestation with respect to energy. It will be defined as the period of uninterrupted execution of the system without the need for recharging batteries. 

Security of a system can also span over several components such as network, storage, and data. In this paper, we will focus on both network and security of data used in human centric AI enabled critical systems.

\section{Limitations of current techniques}

\subsection{Safety}
The mixed approach of model based safety assurance and experimental analysis have several limitations. Experimental safety analysis is often expensive and hence can only be performed on a representative set of scenarios. It is an extremely difficult and often subjective task to select such a representative set that covers an exhaustive set of use cases that may occur in practice. The solution is to supplement experimental analysis with model driven safety verification. It typically involves using a model to estimate the behavior of the system through the use of mathematical algorithms and simulation. The outcome is a set of parameter variations over time starting from an initial condition, a characteristic of the use case. This is often referred to as ``execution" of the system. These executions can then be compared with the safety condition to evaluate the safety of the system. The advantage of this approach is that simulations or mathematical estimations are less expensive and faster. In addition the system can be analyzed for a set of initial conditions (potentially containing infinite use cases) at one go instead of iterating through them one by one. 

This approach has been extensively used for many AI enabled human computer systems such as closed loop blood glucose control, and autonomous cars. However, the general problem of model based safety verification is intractable and cannot be solved accurately in limited time. Researchers have used several methods to approximate the system behavior over time and derive what is often referred to as \textit{reach set}. It is the approximate set of executions of a system for a given set of initial conditions representing use cases. Since the reach set is an approximation, this implies that the estimated behavior of the system for certain use cases are uncertain. Thus if such use cases actually occur in practice then the system behavior can possibly result in unsafe conditions. The usual practice is to design the system such that the probability of occurrence of an unsafe condition due to uncertainty in verification result is minimized. However, long term usage entails that even if the probability is low, still there is a possibility of actual occurrence of an uncertain use case leading to safety violations. These use cases are often referred to as \textit{corner cases} in recent literature. 

There is a new wave of research on \textit{corner cases} including determination of corner cases from execution traces of the system, accurate estimation of execution of a system for corner cases, and system design for avoiding safety violation for corner cases. However, there are significant challenges in each of these areas which are not addressed in recent research.

\subsection{Sustainability}
Although wearable sensors are low power devices, a quick survey on several critical systems such as cardiac monitoring devices, or closed loop blood glucose control systems or gesture recognition systems show that the lifetime of wearables is of the order of weeks if not days. For example, the state of the art electrocardiogram wearable sensors can only sense continuous samples of ECG for a period of a week, continuous glucose monitors (CGM) sensors expire after a week, and in case of gesture based interfaces, the sensors only last for a day. Often such limitations are not due to lack of a power source, as in the case of CGMs, but barring some exceptions availability of energy is one of the key factors. Such a constraint can render long term usage of sensing systems for critical applications difficult and often inconsequential. We explain this using the example of cardiac monitoring.    

The current state of the art of wearable medical monitoring systems, such as ECG monitors, have a lifetime of only about 300 hours i.e., less than 13 days using techniques such as compressive sensing~\cite{zhang2013compressed}. There are several research initiatives to build battery-less sensors which operate using energy scavenged from body heat. However, they can only provide small amount of energy which may not sustain data transmission to a central data repository and hence limit functionality. The state of the art is the insertable cardiac monitor (ICM), which is an event monitor implanted under the skin. The ICM can monitor AF events, and store ECG data related to the AF event. The data stored by the ICM can be reported to a base station on a daily basis. The ICM installation is involves an invasive procedure and is expensive~\cite{bhangu2016long}. Although the procedure lasts for nearly 30 minutes it involves making an incision on the patient’s chest and insertion of the device under the chest bone. The ICM can operate for 3 years without recharge and can detect AF episodes lasting at least 2 mins~\cite{ICM}. The disadvantage of using ICM is that the procedure and device is costly and invasive.

\subsection{Security}
There are several aspects of data provenance problem in wearable sensor driven  AI enabled human centric critical systems. In this paper, we will only focus on the aspect of data manipulation at the source. We will consider the example of medical systems to discuss the data manipulation aspect of data provenance. {\bf Data manipulation attacks} on wearable medical systems (WMS) are attacks that allow adversaries to report incorrect patient health state information.

Both the sensors and the base station are susceptible to data manipulation attacks. Data manipulation attacks on sensors can be mounted in many ways. For instance, by {\em exploiting the open wireless communication channel} between sensors and the base station, as identified for pacemakers \cite{Halperin*08} and insulin pumps \cite{Li*11}. Adversaries can also exploit {\em sensory-channel vulnerabilities}, which involve interfering with the transducers of the sensors and introducing arbitrary sensor measurements into the system. This can be performed using a variety of stimuli including electromagnetic induction \cite{scream}, light \cite{Park*16}, and acoustic waves \cite{Dean*07}. Such sensory-channel attacks can not only be used to tamper with the sensor measurements \cite{GhostTalk}, but also enable arbitrary code execution under specific conditions, as we ourselves recently identified  \cite{scream}. Adversaries can also install malware on the sensors and the base station during routine software updates.

WMS are very different from traditional computing domain in several important ways. Individual sensors and the base station in a WMS are considered medical devices. As patient safety is given a lot of importance with the operation of WMS, the sensors and the base station in the WMS need rigorous device certification from agencies like the Food and Drug Administration (FDA). Moreover, the WMS sensors and base station (which are not a commodity device such as smartphone but rather a specially-designed  device that is designed suitable for medical data processing) themselves have limited computational capabilities. 
Consequently, traditional security approach are just not sufficient to deal with security problems in the WMS context. For instance:

\underline{{\em Communication encryption-based solutions are not sufficient.}} With these new class of attacks, the source (sensors) or intermediate destination (base station) of the data-flow within the WMS itself is compromised. No amount of securing the communication in the middle is therefore going to protect against patient harm.

\underline{{\em Patching may not be without risk.}} Medical devices operate under very strict regulatory regime monitored by the FDA. The 2017 FDA post market guidance allows medical device manufacturers to patch vulnerabilities in their devices without seeking FDA approvals as long as the patch does not change the fundamental functionality of the device. However, the patching infrastructure is fraught with risks. Medical devices have also been compromised by leveraging the fact that they do not typically authenticate the received software and libraries during on-field updates. A compromise of the manufacturer's servers can thus be used to compromise sensors and base station during patching. 

\underline{{\em Malware detection solutions are not easy to introduce on WMS.}}  WMS sensors and base station are extremely heterogeneous and often built using very constrained embedded platforms. This makes deployment of general-purpose anti-virus (AV) software infeasible without incurring extensive customization costs \cite{WattsupDoc}. Further, even if AV software is present, in many cases manufacturers switch them off during firmware updates \cite{WattsupDoc}. 

\underline{{\em Hardware-oriented solutions are expensive.}} Researchers have proposed upgrading the device hardware to prevent some of the exploits using techniques such as tamper-proof hardware, device shielding, or advanced filtering systems \cite{GhostTalk}. These approaches increase the cost, weight, and energy consumption of the devices, thus affecting the wearability of the device and mobility of the patient.

\section{Personalized Model based Solutions}
In our research, we explore an interesting theme of using personalized models of the underlying processes to solve the challenges of safety, security, and sustainability for long term practical deployments. The idea is to develop a parameterized model of the underlying physical process that is updated over time with new sensor observations. Such parameterized model is used in safety analysis, sustainability adjustments and security protocols. We consider the example of medical control systems and demonstrate the usage of personalized model for safety, sustainability, and security assured design of AI enabled human centric control systems. 

\subsection{Safety}
Personalized model can be utilized for corner case analysis in safety verification of AI enabled human centric critical systems. Models can be used in two aspects: a) usage model of the system, and b) model of the physical processes that influence the operation of the system in real time. The first type of model is typically discrete and probabilistic while the second type of model has to be continuous. 

Usage model of the system depends on the unique contexts presented by the user. This can be influenced by several factors including movement, physiological states, mental states, and behavioral patterns such as sleep, eating, physical activity. As an example, if we consider the closed loop blood glucose control system, physical activity plays an important role in changing the underlying glucose insulin dynamics. This may result in unique corner cases that may no be analyzed in a controlled experimental setting. A specific scenario of interest is performing physical activity immediately after taking a meal. The prescribed usage of an AP system is to report the meal amount and use a Bolus Wizard to obtain a certain dosage of insulin. However, if a person performs extensive physical activity immediately after taking a meal, the insulin sensitivity increases significantly. Hence taking in the prescribed amount of insulin can potentially lead to dangerous hypoglycemic events. This corner case is entirely dependent on the behavioral pattern of the user. However, such usage pattern changes frequently because most subjects do not have a fixed schedule of physical activity. Hence incorporating such effects in model based verification requires testing of a significantly large number of action sequences which is computationally expensive. However, a Markov chain based model of the user's exercise pattern can narrow down the use cases to be tested in simulation. By learning the probabilities from the usage data of the system the Markov chain based model can provide us highly probable corner cases, which can then be simulated to obtain safety results. This idea has been explored in some details on a mobile healthcare setting for safety analysis of medical control systems~\cite{banerjee2015analysis}. The Markov chain model can be further guided by expert rules. The usage data can be used to characterize these rules as rare or common. Through such characterizations rare rules can be preferred to determine corner cases that occur infrequently but are critical to the safety analysis. 

Continuous models of underlying processes in the system are typically deterministic in nature. For example, the underlying process in an AP system is the glucose insulin dynamics, which is typically modeled using differential equations. The insulin sensitivity is one coefficient in the equations that affects the overall dynamics. This insulin sensitivity is affected by physical activity. Using mining techniques we can change the differential equation model of the glucose insulin dynamics to reflect the dynamic changes in insulin sensitivity. Hybrid model mining from input output traces is a new area of research that allows us to derive dynamic models of the user~\cite{lamrani2018hymn}. This enables usage of time varying models of the system in simulation. 

Time variance is an inherent property of human physiological processes. Time variance can lead to contexts that may not be observed in testing but observed in practice. The model mining technology enables modeling of time variance in silico. This enables discovery of new corner cases in simulation that may not have been observed in testing time with time invariant simulation techniques.

\subsection{Security}
Personalized models can also be used to detect data manipulation attacks on WMS, and thus determining the trustworthiness
of the information collected by them. The approach is to encode the unique interrelationship between the measured physiological
signals measured by a WMS using a machine-learning model. This allows us to capture the
characteristics of the patient’s physiological processes. Consequently, introduction of incorrect patient data
by adversaries (in the form of fake physiological signals) will not match the known characteristics of the patient’s
physiological processes and will be detected by our model. For example, electrocardiogram (ECG),
blood pressure, and plethysmogram are distinct vitals emanating from the underlying cardiac process of the
patient. Any erratic change in the relationship of the physiological signals, from a malicious manipulation of
the ECG or blood pressure or pulse-ox measurements can indicate data manipulation.
We take a physiology-driven approach because in the context of WMS we cannot rely on redundant
devices that measure the same physiological signal to detect data manipulation attacks. WMS typically
have only data-source of a type for usability reasons. Although there exists significant body of research
on generic models of physiological processes, the proposed approach has to be patient-specific to
account for the individual variations in a free living scenario. We expect our data-driven system for detecting
data manipulation to execute on the cloud outside the WMS. It will then vet every stream of data coming
from a patient’s WMS at run-time and allow it to be stored in the patient’s EHR only if it is deemed to be not
manipulated.

Model based approach can also be taken to perform end to end security. Physiological value based key agreement (PSKA) has been proposed by the authors that uses the human bodily signals to agree on a secret between different wearable sensors~\cite{venkatasubramanian2010pska}. However, this idea can be combined with generative models of signals to derive a model based security methodology. The principle is to execute PKA thus enabling authenticated key agreement but replace the raw physiological signals on one end with the diagnostically-equivalent synthetic signals obtained from a trained physiological generative model. This enables model based encryption. The generative model is a partially time variant model. It has a morphological component that remain more or less invariant over the lifetime of a user, but it has a time varying component that varies fast. For example, for electrocardiogram signals, the morphology is the beat shape and the time varying parameters are the heart rate related features. Hence usage of personalized model enables time varying security measures. This is especially important because it increases the effort required from an adversary to break the security measure.

\subsection{Sustainability}

Compression techniques that enable sensing at lower frequencies than theoretically required, and saving communication through the usage of signal models can potentially facilitate continuous cardiac monitoring during free living conditions. This is because by reducing the sample size, they reduce the execution time of the data processing algorithms. 

There are two competitive compression techniques: a) compressive sensing (CS) that allows accurate recovery of signals with fewer samples than Nyquist rate~\cite{cs6}, and b) generated model based resource efficient monitoring (GeMREM), which compares a signal with a pre-learned model and reduces data transmission if signal matches model~\cite{gemrem}. On one hand CS provides sensing reduction but has a complex recovery method, which makes it inefficient with respect to energy and storage requirements when implemented in a smartwatch or a smartphone. On the other hand, GeMREM provides no sensing reduction and requires the sensor to process the data and match with a model, but it gives orders of magnitude more communication reduction than CS and has a very simple recovery algorithm. The lightweight recovery algorithm enables resource efficient execution of GeMREM in a smartwatch or smartphone.

The notion of CS and GeMREM can be potentially combined to develop a novel generative model based compressive sensing (GenCS), which provides high order of sensing compression with simple sensors, and resource efficient recovery. 

In their previous work~\cite{gemrem}, the authors have shown the usage of GeMREM on ECG signals. Extensive testing on data from the MIT BIH database on 20 subjects shows an average compression ratio of 40 for GeMREM as opposed to the theoretical bound of 2.5 seen for CS. 
The shape of a beat makes ECG signal a non-sparse signal not only in time domain but also in DWT, DFT, and DCT domains. However, the temporal parameters are only related to the R peaks. Hence a signal with only R peaks can be approximated much more accurately using greater sparsity. GenCS senses the signal to only recover the temporal parameters (R peaks for ECG) and suppress the morphological parameters. The morphological parameters can be learned from a signal snippet sensed at the Nyquivst rate. On obtaining the temporal parameters, the entire signal can be recovered by combining the morphological and temporal parameters.     

\begin{figure}
	\centering
	\includegraphics[width=0.9\columnwidth]{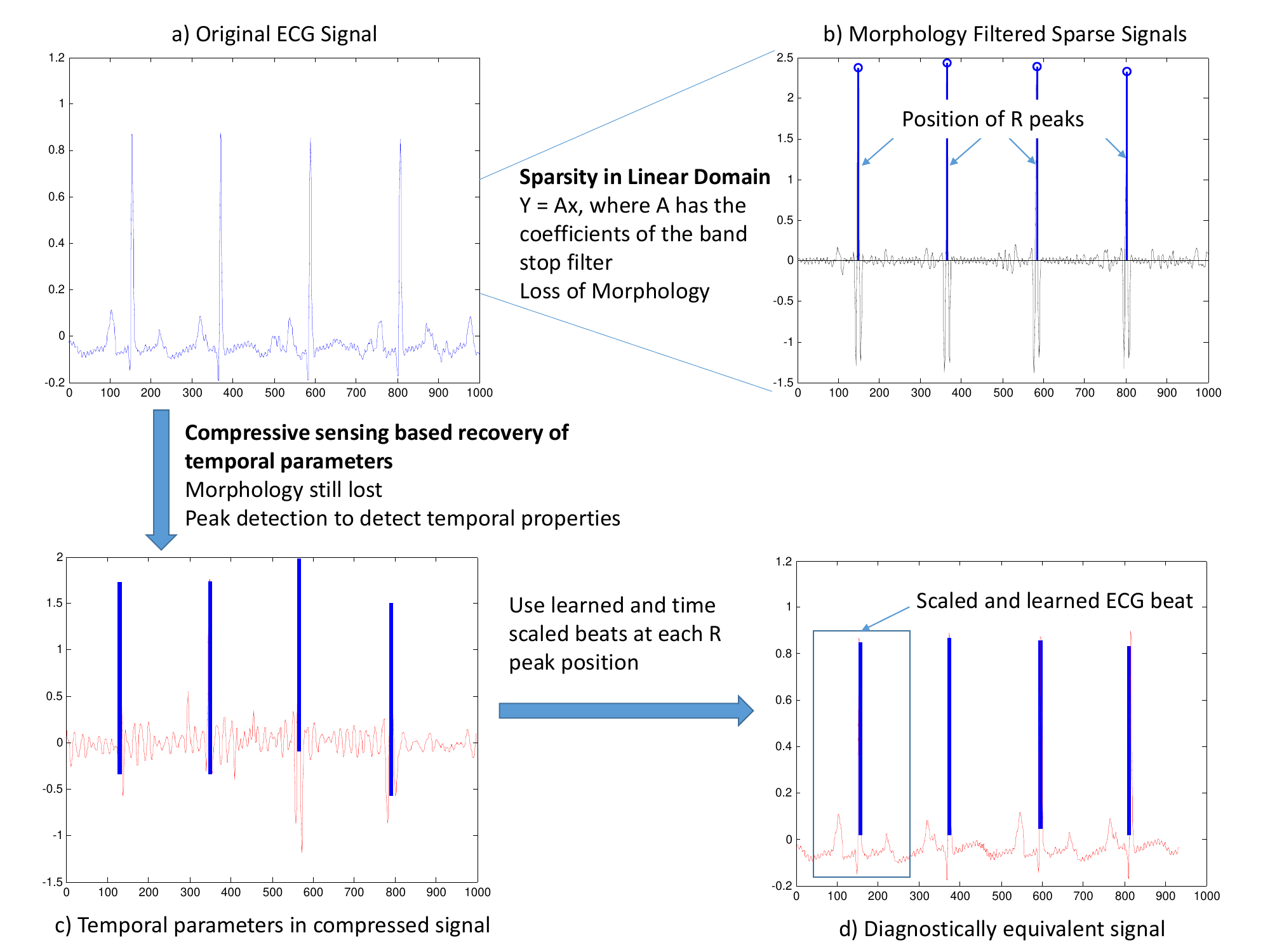}
	\caption{Example execution of GenCS on ECG data.}\label{fig:ECGFigfilt}
\end{figure}

For ECG signals the shape characteristics can be suppressed using a low pass and high pass filter combination as discussed in~\cite{gemrem}. A digital filter (Equation \ref{eqn:filt}), at low pass cut-off frequency of 5 Hz and high pass cutoff at 12 Hz, can effectively eliminate the P, Q, S, and T waves, and only keep the R peaks in the signal. 
\begin{eqnarray}
\label{eqn:filt}
\scriptsize y[i] = 2y[i-1] - y[i-2] + x[i] -2x[i-6] + x[i-12], \\\nonumber
\scriptsize z[i] = 32x[i-16] - z[i-1] + x[i] - x[i-32],
\end{eqnarray}
The resulting signal $z$ as shown in Figure \ref{fig:ECGFigfilt} b has only the temporal parameters and the morphology is suppressed. The signal $z$ is compressible to a sparse vector and can be recovered with very less number of samples. Figure \ref{fig:ECGFigfilt} c, shows the GenCS recovered signal from the original signal. The sensing matrix $\Phi$ used for this purpose was generated using a Bernoulli distribution. The transformation matrix $\Theta$ for making the original signal sparse was obtained by first converting the band-stop filter in Equation \ref{eqn:filt} into a matrix form and multiplying it with the DWT matrix. The morphological parameters of the signal were learned previously using the curve fitting technique. The ECG beat shape is then centered at each R peak obtained from the signal in Figure \ref{fig:ECGFigfilt} and temporally scaled to match the heart rate~\cite{gemrem} and obtain a diagnostically equivalent signal.

Clinical data from 69 subjects collected in various settings including cardiac ICU, free living condition, and neonatal ICU shows the following:

\noindent{\bf -} GenCS can improve the compression ratio of CS by 5 fold for recovering a signal that is diagnostically equivalent (as confirmed by a cardiac surgeon) to the original signal.\\
\noindent{\bf -} GenCS recovery algorithm has nearly 2.7 times less energy consumption than CS when implemented in a smartphone. Hence the lifetime of a smartwatch or smartphone will be improved nearly 3 times when GenCS is selected as the compression scheme.\\
\noindent{\bf -} To recover 2 seconds of electrocardiogram (ECG) data in a smartwatch using CS it took 2.4 s. This means that before recovering the ECG data the sensor buffer will be full and overflow resulting in loss of data. Hence, CS can only be implemented for a limited amount of time in a smartwatch. However, the same recovery takes 1.2 s for GenCS. Thus, GenCS enables unlimited data recovery using a smartwatch.   

\begin{figure}
	\centering
	\includegraphics[width=0.9\columnwidth,clip=true,trim=0 100 0 0]{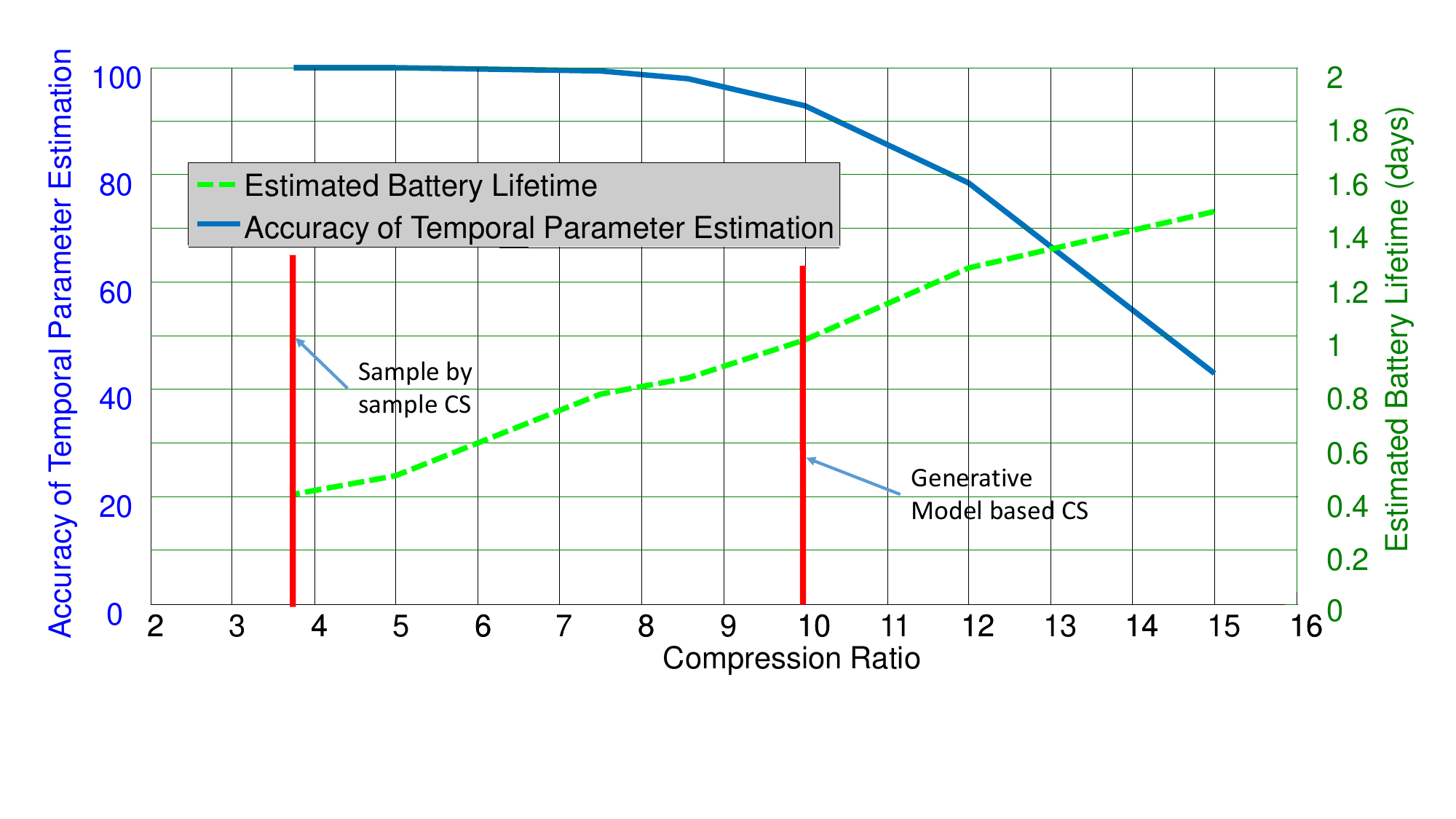}
	\caption{Accuracy and lifetime comparison for generative model based CS and sample-by-sample CS.}
	\label{fig:Lifetime}
\end{figure}

\noindent{\bf Energy consumption and lifetime:} To compute the lifetime of smartphone due to the execution of the recovery algorithm, we consider that there are no other apps running. Hence, the whole battery with a capacity of 1625 mAh is utilized by the recovery algorithm. Figure \ref{fig:Lifetime}, shows the lifetime assuming a linear co-relation between energy consumption and battery discharge rate. The lifetime is shown with respect to compression ratio of the CS techniques. In the same graph, we also show the accuracy of extraction of temporal parameters. Figure \ref{fig:Lifetime} shows that as compression ratio increases, the lifetime also increases however, the accuracy decreases. For the GenCS technique, the energy consumption is nearly three times lower than the CS recovery algorithm. The smartphone lifetime for the GenCS for a diagnostic accuracy of 10\% is around 1 day which is nearly 15 hrs more than the lifetime of the CS.

\section{Conclusion}
Personalized models are seen to solve several important problems related to safety, security and sustainability. Extraction of corner cases and time varying simulation to derive rarest of failures in AI enabled human centric critical systems are important problems which can be potentially solved using personalized models. Data driven learned model based detection of data manipulation attacks is a novel idea that is explored by recent researchers and is also dependent on the personal characteristics of a person. Generative model based encryption takes a different dimension that automatically allows for key refreshment because of frequent changes of physiological states. Personalized models can also be used in several energy efficient sensing techniques that are being explored in recent research. This shows that usage of personalized models is an interesting prospectus in solving important problems for safety, sustainability and security assured development of AI enabled human centric critical systems and should be explored in depth.
\ifCLASSOPTIONcompsoc
\else
\fi

\ifCLASSOPTIONcaptionsoff
  \newpage
\fi



\balance
\scriptsize
\bibliographystyle{IEEEtran}
%
\bibliography{WirelessHealth2016bib,NSFCCF,Fengbo_Ref,Ref} 
\end{document}